\def\year{2019}\relax
\documentclass[letterpaper]{article} 
\usepackage{aaai19}  
\usepackage{times}  
\usepackage{helvet}  
\usepackage{courier}  
\usepackage{url}  
\usepackage{graphicx}  
\usepackage{amsmath}
\usepackage{amssymb}
\usepackage{breqn}
\usepackage{dirtytalk}
\usepackage{booktabs}
\usepackage[usenames, dvipsnames]{color}

\begin{document}
%
\title{A Layer-Based Sequential Framework for Scene Generation with GANs
}
\author{Mehmet Ozgur Turkoglu\textsuperscript{1}, William Thong\textsuperscript{2}, Luuk Spreeuwers\textsuperscript{1}, Berkay Kicanaoglu\textsuperscript{2}\\
\textsuperscript{1}University of Twente, \textsuperscript{2}University of Amsterdam\\
moturkoglu@gmail.com, l.j.spreeuwers@utwente.nl, \{w.e.thong, b.kicanaoglu\}@uva.nl\\
}

\maketitle
\begin{abstract}
The visual world we sense, interpret and interact everyday is a complex composition of interleaved physical entities. Therefore, it is a very challenging task to generate vivid scenes of similar complexity using computers. In this work, we present a scene generation framework based on Generative Adversarial Networks (GANs) to sequentially compose a scene, breaking down the underlying problem into smaller ones. Different than the existing approaches, our framework offers an explicit control over the elements of a scene through separate background and foreground generators. Starting with an initially generated background, foreground objects then populate the scene one-by-one in a sequential manner. Via quantitative and qualitative experiments on a subset of the MS-COCO dataset,  we show that our proposed framework produces not only more diverse images but also copes better with affine transformations and occlusion artifacts of foreground objects than its counterparts.

\end{abstract}

\section{Introduction}

The visual world we sense, interpret and interact everyday is a complex composition of interleaved physical entities.
Scenes we perceive can then be altered due to accidental factors such as \textit{occlusions} caused by inter-entity interactions and \textit{self-occlusions} caused by differing vantage points with respect to the camera.
These factors do not only make the problem harder but also diversifies the possible space of configurations given a fixed set of observable entities. Consequently, scene generation is challenged by high sample complexity in addition to photorealism constraints.

\begin{figure}[t!]
\centering
\includegraphics[width=8.0cm]{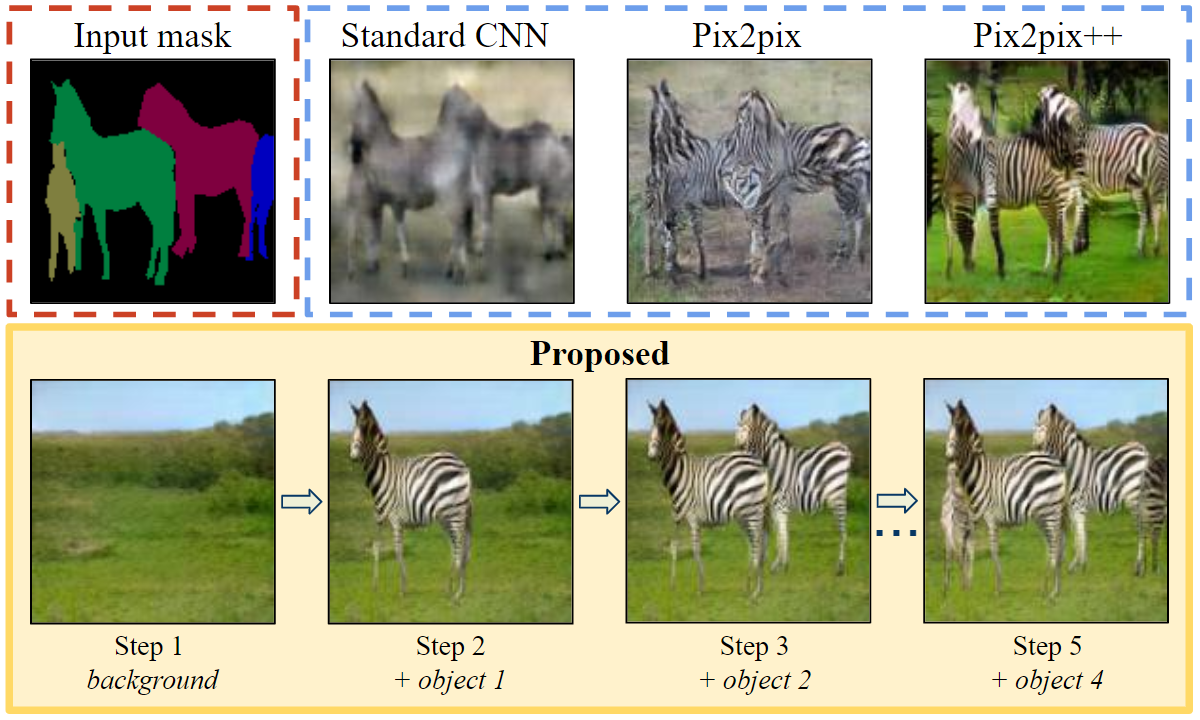}
\caption{The proposed image generation process. Given a semantic layout map, our model composes the scene step-by-step. The first row shows the input semantic map and images generated by state-of-the-art baselines.}
\label{intro}
\end{figure}

Scene generation problem has been studied extensively in \cite{hertzmann2001image,tappan2008knowledge,chen2009sketch2photo,isola2013scene}, however the interest has certainly expanded with the advent of Generative Adversarial Networks (GANs)~\cite{gan}. Although recently there have been great strides in the quality and resolution of generated images~\cite{spec_norm,pix2pixHD}, it has mostly been addressed as learning a mapping from a \textit{single} \textit{source}, \textit{e.g.} noise or semantic map, to \textit{target}, \textit{e.g.} images of giraffes, discarding almost all of the underlying complexity that forms a scene. This formulation sets two major restrictions on \textit{(i)} ability to control scene elements, \textit{e.g.}~\say{What objects will appear?} or~\say{How will they appear?}, \textit{(ii)} ability to manipulate an existing scene, \textit{e.g.}~\say{Make that cow larger!}, while keeping rest of the scene unaltered. One possible explanation is that a single \textit{source} representation such as a noise vector $z$
is insufficient to explain all the complexity involved in describing a scene meaningfully. 

In this paper, we tackle the scene composition task with  a layered structure modeling approach. This formulation helps reducing the complexity of the problem since we deal with one subproblem at a time. Furthermore, our framework allows for explicit control mechanisms at multiple levels such as scene layout and individual object appearances which lack in the previous approaches. 

Our main idea resembles how a landscape painter would first sketch out the overall structure, \emph{e.g.} first a background such as mountain ranges or rivers, and later embellish the scene gradually with other elements to populate the scene, \emph{e.g.} foreground instances such as trees and animals. Our proposed sequential framework works in an analogous manner to the procedure as depicted in Figure~\ref{intro}. It starts with a background generator to fill the canvas at the first time step. In every consecutive time step 
a foreground generator draws then a new entities conditioned on an instance mask and the canvas passed on from previous time step giving the user control over \textit{where} and \textit{how} the object will appear in the scene. Our model simultaneously aims to preserve the previously drawn background scene and added entities.

We propose a novel  image generation framework to sequentially compose a scene, element-by-element. We achieve this by an explicit control over the elements of a scene through separate background and foreground generators. This allows user control over the objects to generate, as well as, their category, their location, their shape and their appearance. Moreover, the sequential composition better handles affine transformations of objects and better copes with object occlusions than existing conditional GAN models. Experiments are carried out on a subset of the MS-COCO dataset. Both qualitative and quantitative results show the strength and the advantages of the proposed sequential framework for scene composition compared to state-of-the-art image generation baselines.


\section{Related work}

\textbf{Conditional Image Generation.} Various approaches have been proposed to control the generation process by conditioning on a class label~\cite{acgan}, an attribute vector~\cite{attribute_controller,ding2017exprgan}, a text description~\cite{reed,feifei}, or a semantic map~\cite{pix2pix,google_paper}.~\citeauthor{what_where} learn to control the foreground object location by conditioning on bounding boxes and keypoint coordinates.~\citeauthor{pix2pix} propose conditional adversarial networks as a general purpose solution to image-to-image translation problems. Their model can generate a scene from a semantic layout map similar to our work. Nevertheless, the image diversity and the limited control over scene elements are limited.

~ \citeauthor{pix2pixHD} address these limitations of previous conditional models and propose a feature-embedding approach using additional encoder network in order to improve the diversity and controllability.  Similar to our method, their method allows an element-level control; however, they use pre-defined low-dimensional feature map to control the appearance of the elements which may limit the diversity.

~\citeauthor{mcgan} introduce a new method to learn to generate a foreground object image by conditioning on both the text description, the foreground object mask, and the given background image without altering it. This is the most relevant work to our paper. However, the method can only generate a single foreground object when the background image is given. In contrast, our framework can generate both a background and an arbitrary number of foreground objects. 

\noindent\textbf{Sequential Image Generation.} 
A coarse-to-fine generation process can be cast as a sequential image generation process.
~\citeauthor{lapgan} introduce a sequential model with a cascade of generative models, each of them captures the image structure at a particular scale of a Laplacian pyramid.~\citeauthor{stackGAN} improve the image quality by increasing image resolution with a two-stage GANs.~\citeauthor{rBTN} generate a face image from a small face patch iteratively. They refine the image by transformations between two image manifolds. These approaches differ significantly with our method but we share the same insights to generate an image sequentially by decomposing the generation process in smaller subtasks.

Similar to our method,~\citeauthor{stgan} and~\citeauthor{lrgan} adopt a layered structure modeling and compose a scene sequentially. However,~\citeauthor{stgan} focus on finding the appropriate geometric transformation of an object in a background layout, in an iterative manner. In our work, the user defines the geometry of the object and no transformation is involved. Their objective is then orthogonal to ours. \citeauthor{lrgan} focus on the unsupervised discovery of object masks, in a layer-wise process. Contrary to our work, the foreground generator produces a foreground mask instead of being conditioned on it. Such a scheme does not provide the same level of controllability and flexibility as ours. Moreover, their experiments were restricted to one object instance per image compared to 2.61 instances per image on average in our case.

\section{Methodology}

\subsection{Problem formulation}

The objective is to compose a realistic scene while allowing element-level control. In this work, a scene is defined as a composition of background and foreground elements. At training time, we are given a set of images $x \in \mathbb{R}^{H\times W\times 3}$'s and associated semantic layout maps $M$'s drawn from a joint probability distribution $p(x,M)$. Our goal is to learn a function $\Phi: M\rightarrow x$ which maps a semantic layout map to a scene image. $\Phi$ models the conditional probability distribution $p(x|M)$. In our formulation, a semantic layout map $M$ consists of a set of foreground object masks $M=\{M_1,\ldots,M_T\}$ where $T$ is a number of foreground objects in the scene and $M_t \in \{0,1\}^{H\times W\times N}$\ is a binary tensor that defines the $t^{th}$ foreground object location, size, shape, and class with $N$ being the number of object categories in the dataset. 
For convenience, we define two additional variables: (1) an aggregated semantic map $M_{agg}=\max_{t \in \{1\ldots T\}} M_t \in \{0,1\}^{H\times W\times N}$ that is obtained by taking the maximum value for each element in the set and (2) an occupancy map $m_{t}^{occ} \in \{0,1\}^{H\times W}$ ($m_{t(ij)}^{occ}=\max_{n \in \{1\ldots N\}} M_{t(ijn)}$) whose element is 1 if the corresponding pixel is occupied by a foreground object. 

At both training and inference steps, the scene generation process is then conditioned on both $M$ and a collection of noise vectors $z$. A new noise vector is sampled from a normal distribution for every step of the generation process, which gives individual control over the appearance of the generated elements.

\begin{figure}[t]
\centering
\includegraphics[width=7.5cm]{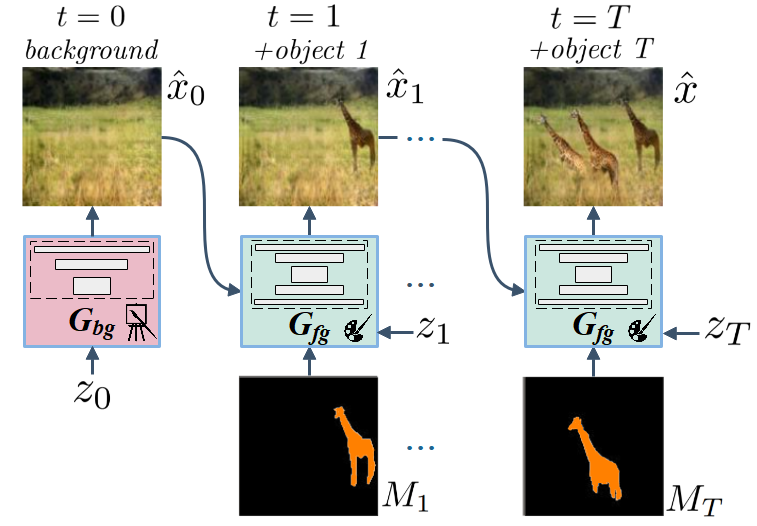}
\caption{The proposed framework overview. $G_{bg}$, $G_{fg}$ are the background and foreground generators, respectively.}
\label{generation_process}
\end{figure}

Consider Figure~\ref{generation_process}. The main objective is broken down into two simpler sub-tasks. First, we generate the background canvas $\hat{x}_0$ with the background generator $G_{bg}$ conditioned on a noise $z_0$. Second, we sequentially add foreground objects with the foreground generator $G_{fg}$ to reach the final image $\hat{x}$ ($=\hat{x}_T$), which contains the intended $T$ foreground objects on the canvas.

\subsubsection{Background generator}
The background model generates a background image without any foreground objects, similar to background distribution of the training images. The architecture consists of two main branches: (1) the first branch generates a background image $\hat{x}_{0}$ from a noise vector  $z_0$; (2) the second branch generates a scene denoted by $\hat{x}'_{0}$ from both a semantic map $M$ and the intermediate features of the first branch. The second branch is then similar to an image translation problem, except that it is also conditioned on the features from the first branch. At inference time, we discard the second branch and only keep the first branch.

For training, we define an interleaved loss function for each branch: (1) A reconstruction loss $L_{rec}$ measures the L2 distance between a generated scene $\hat{x}'_{0}$ and its generated background $\hat{x}_{0}$. $L_{rec}$ only penalizes pixels outside of the foreground masks. 
\begin{dmath}
L_{rec}  =  \mathbb{E}_{z_0,M}[||(1-m_{agg}^{occ})\odot(\hat{x}_{0}-\hat{x}'_{0})||_2]
\end{dmath}
$\odot$ is a pixel-wise multiplication and similar to $m_{t}^{occ}$, $m_{agg}^{occ}$ is an occupancy map obtained from $M_{agg}$ ($m_{agg(ij)}^{occ}=\max_{n \in \{1\ldots N\}} M_{agg(ijn)}$).  (2) An adversarial loss $L_{adv}$ encourages fake $\hat{x}'_{0}$ images to look like real images $x$. By using these two losses together, the discriminator implicitly considers pixels outside of the foreground mask and this formulation gives freedom to implicitly inpaint the foreground regions in $\hat{x}_{0}$.

\small
\begin{dmath}
 L_{adv} = \mathbb{E}_{(x,M)}[\log D(x,M_{agg})] + \mathbb{E}_{z_0,M}[\log(1-D(\hat{x}'_{0},M_{agg})]
 \label{eq:lossBG_ADV}
\end{dmath}
\normalsize
Additionally, we apply a feature matching loss $L_{fm}$ on $\hat{x}'_{0}$ to stabilize the training~\cite{imp_tech}. So the overall objective is the following:
\begin{dmath}
L = L_{adv} + \lambda_{r} L_{rec} + \lambda_{fm} L_{fm}
\label{eq:lossBG}
\end{dmath}
where $\lambda$'s are trade-off parameters set empirically.

\begin{figure}[t]
\centering
\includegraphics[width=8cm]{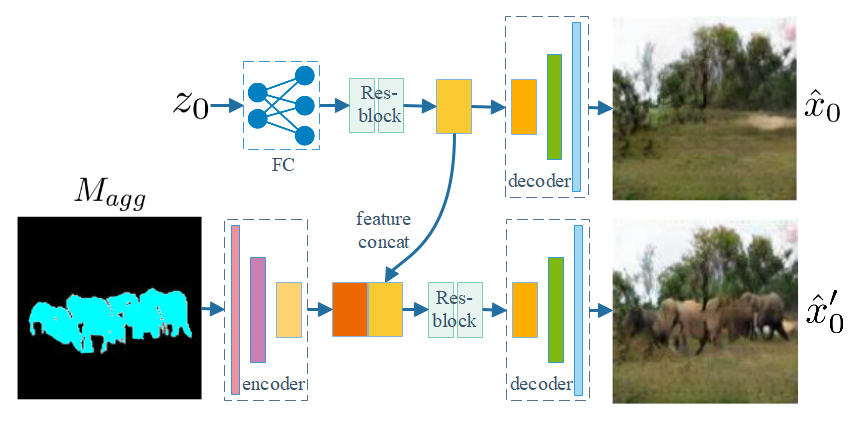}
\caption{The background model network architecture. The $G_{bg}$ generates two images: (1) the background image, $\hat{x}_{0}$ and (2) the scene image, $\hat{x}'_{0}$ that is conditioned on a semantic layout map, $M_{agg}$.}
\label{background_model}
\end{figure}

\subsubsection{Foreground generator}

\begin{figure}[t]
\centering
\includegraphics[width=8cm]{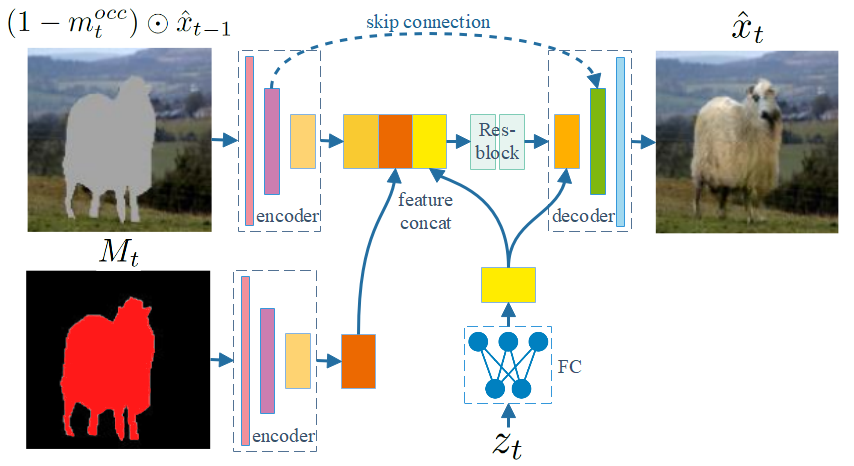}
\caption{The foreground model network architecture. The $G_{fg}$ encodes the given object semantic structure, $M_t$ and the previously generated scene $\hat{x}_{t-1}$ using separate pathways and generates a new image, $\hat{x}_{t}$.}
\label{foreground_model}
\end{figure}

The foreground generator draws elements of a scene one-by-one. It modifies the canvas within a region or a subset of pixels. Its design makes it aware of the previously generated scene in terms of semantic content and global image illumination. It takes as input the previously generated scene $\hat{x}_{t-1}$, the current foreground object mask $M_t$, and a noise $z_t$. We illustrate the model in Figure \ref{foreground_model}. The action of $G_{fg}$ over the scene $\hat{x}_{t-1}$ at time step $t$ can be expressed as follows:

\begin{dmath}
\hat{x}_{t} = G_{fg}((1-m_t^{occ})\odot \hat{x}_{t-1} , M_t , z_t)
\end{dmath}
In this formulation, the foreground object generated by $G_{fg}$ is conditioned on the previous scene ($\hat{x}_{t-1}$) ignoring the regions $M_t$ hides. This formulation allows us to approach object generation as an image inpainting problem. As a result the foreground model can be trained similar to GANs-based image inpainting models~\cite{impaint_1,impaint_2} instead of recurrent training. During training, real images are used as an input to the generator. For each forward pass, one foreground element within the selected image is randomly picked and used.
$G_{fg}$ is trained in an adversarial scheme. Two discriminators $D_{global}$ and $D_{local}$ are jointly trained with $G_{fg}$. $D_{global}$ encourages a global image realism: 

\small
\begin{dmath}
L_{global} = \mathbb{E}_{(x,M)}[\log D_{global}(x,M_{agg})] + \mathbb{E}_{z_t,(x,M),M_t}[\log (1-D_{global}(G_{fg}((1-m_t^{occ})\odot x,M_t,z_t),M_{agg}))]
\label{eq:lossGlobal}
\end{dmath}

\noindent while $D_{local}$ encourages the foreground object realism by focusing on the foreground region:

\begin{dmath}
L_{local} = \mathbb{E}_{(x,M)}[\log D_{local}(\mathcal{I}(x),\mathcal{I}(M_{agg}))] + \mathbb{E}_{z_t,(x,M),M_t}[\log (1-D_{local}(\mathcal{I}(G_{fg}((1-m_t^{occ})\odot x,M_t,z_t)),\mathcal{I}(M_{agg})))]
\label{eq:lossLocal}
\end{dmath}
\normalsize
\noindent
where $\mathcal{I}$ denotes fully-differentiable bilinear interpolation operation that crops the region of interest (object bounding box) and scale it to original spatial sizes. 
The third loss defined in Eq. (\ref{eq:lossL2}) encourages the generator to reconstruct the input image outside the bounding-box, $b_t^{occ} \in \{0,1\}^{H\times W}$. The idea for using a bounding-box instead of a mask, $m_t^{occ}$ is to give the generator, $G_{fg}$ more flexibility to modify the surrounding of the object accordingly.
\small
\begin{dmath}
L_{rec}  =  \mathbb{E}_{z_t,x,M_t}[||(1-b_t^{occ})\odot(G_{fg}((1-m_t^{occ})\odot x,M_t,z_t)- x)||_2]
\label{eq:lossL2}
\end{dmath}
\normalsize
Also, the feature matching loss $L_{fm}$ is imposed as similar in the background model. The overall objective is following.

\begin{dmath}
L = L_{global} +  \lambda_{l}L_{local} + \lambda_{r} L_{rec} + \lambda_{fm} L_{fm}
\label{eq:lossFG}
\end{dmath}
where $\lambda$'s are trade-off parameters set empirically.

Following~\citeauthor{multimodal}, the noise representation is concatenated at both the output of the encoder and within the decoder to increase stochasticity.
Besides, skip connections~\cite{unet} are added to improve the reconstruction process (see Figure~\ref{foreground_model}).


\section{Experimental setup}
\subsection{Dataset}
MS-COCO dataset~\cite{mscoco} is used to evaluate the performance of our proposed model. The dataset contains 164K training images over 80 semantic classes. Images are annotated with object semantic masks and bounding boxes.
MS-COCO contains images with multiple objects in natural environments and under different viewpoints. To ease the scene composition problem,  six semantically-related categories are chosen: \textit{sheep}, \textit{cow}, \textit{bear}, \textit{elephant}, \textit{giraffe} and \textit{zebra}. These classes have a similar background distribution and there are around 11K images in total for these classes.    

\subsection{Baselines}
We compare our proposed sequential model against three different baselines both quantitatively and qualitatively. All baseline models are non-sequential. They only learn a mapping from the semantic masks to the real image domain. In all experiments, the image size is set to 128x128 pixels.

\subsubsection{Standard CNN}
is a simple baseline without an adversarial loss. It learns a mapping from a semantic layout to a real image with an L1 reconstruction loss. The mapping is deterministic, i.e. the model is memorizing all the training samples. The architecture is similar to Pix2Pix (described below) but without a discriminator.

\subsubsection{Pix2pix}
is the state of the art baseline for the image-to-image translation problem~\cite{pix2pix}. Pix2Pix is designed to learn a mapping from one image domain to another image domain. In our case, the task is to map the semantic layout domain to the real image domain. Similar to a conditional GANs, Pix2Pix is conditioned on a semantic layout. In order to stabilize the training, the L1 reconstruction loss is utilized. Despite the adversarial training, the stochasticity of the generated images is limited. Randomness is provided only in the form of dropout, which is applied on several layers of the generator at both training and test time. Pix2pix is trained exactly as described in~\cite{pix2pix}.

\subsubsection{Pix2pix++}
is an enhanced version of Pix2Pix. Since the publication of Pix2Pix, there has been several improvements for the training of GANs architectures that we adopt in our proposed sequential model. For fair comparisons between Pix2Pix and our method, we apply these additional techniques. Hence, Pix2Pix++ consists in the same discriminator and generator architectures as our proposed model. Moreover, it also integrates a spectral normalization in the discriminator and a feature matching loss.

\begin{figure}[t]
\centering
\includegraphics[width=8cm]{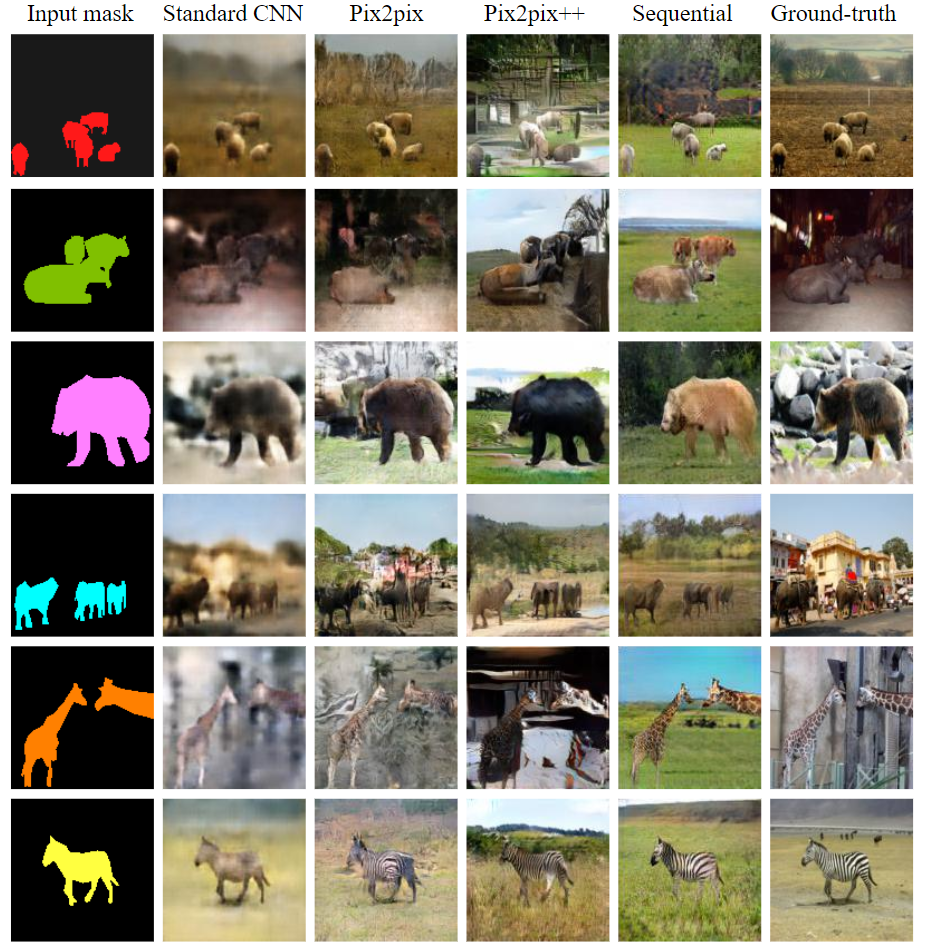}
\caption{Comparison with state of the art models using  object masks from the training set. From top to bottom: \textit{sheep}, \textit{cow}, \textit{bear}, \textit{elephant}, \textit{giraffe} and \textit{zebra}. The ground truth corresponds to to the original image.
}
\label{train}
\end{figure}

\subsection{Evaluation metrics}

\subsubsection{Frechet Inception distance (FID).}
\citeauthor{FID} proposed recently a more consistent metric with human judgment in terms of visual fidelity and with visual disturbances. It measures the Wasserstein-2 distance between the image data distribution of the real images and the generated images in the latent space~\cite{FID}. Features come from the Inception-V3 network~\cite{inceptionV3} pre-trained on the ImageNet dataset (output of the 3rd max-pooling layer). A low FID means that generated image samples are similar to real images in terms of visual quality and semantic content. In our experiments, FID scores are computed based on 10k generated images conditioned on random noise vectors and semantic maps of the training set.

\subsubsection{Semantic segmentation accuracy.}
An indirect way to measure the generated image quality is to perform semantic segmentation with an off-the-shelf model~\cite{pix2pix,pix2pixHD,cycleGAN}. In our experiments, we used Deeplab~\cite{deeplab} pre-trained on the MS-COCO dataset as a segmentation model.
We then measure the Intersection-over-Union (IoU) score, which computes the ratio the intersection over the union of the semantic prediction and ground truth regions.
Similarly to the FID score, we report the mean IoU score based on 10k generated images conditioned on random noise vectors and semantic maps of the training set. Additionally, we also compute the mean IoU scores on the images generated conditioned on semantic maps of the validation set (450 images).

\subsection{Implementation details}
Following~\citeauthor{spec_norm}, both generator and discriminator architectures are based on ResNet. A spectral normalization regularizer is also applied during training. Parameters are updated with the Adam optimizer ($\beta_1=0$, $\beta_2=0.9$, learning rate of $2e^{-4}$ and divided by 2 every 80 epochs)~\cite{adam}. All the models are trained for 480 epochs with a batch size of 16. The parameters of the generators are updated after 5 updates of the discriminator.
The trade-off hyper-parameters in the foreground generator loss function (Eq.~\ref{eq:lossFG}) are set to $\lambda_{l}=0.1$, $\lambda_{r}=1e^{-5}$, $\lambda_{fm}=1$ and in the background generator loss function (Eq.~\ref{eq:lossBG}) to $\lambda_{r}=100$, $\lambda_{fm}=1$. The code is available at 
\url{https://github.com/0zgur0/Seq_Scene_Gen}.

\begin{figure}[t]
\centering
\includegraphics[width=8cm]{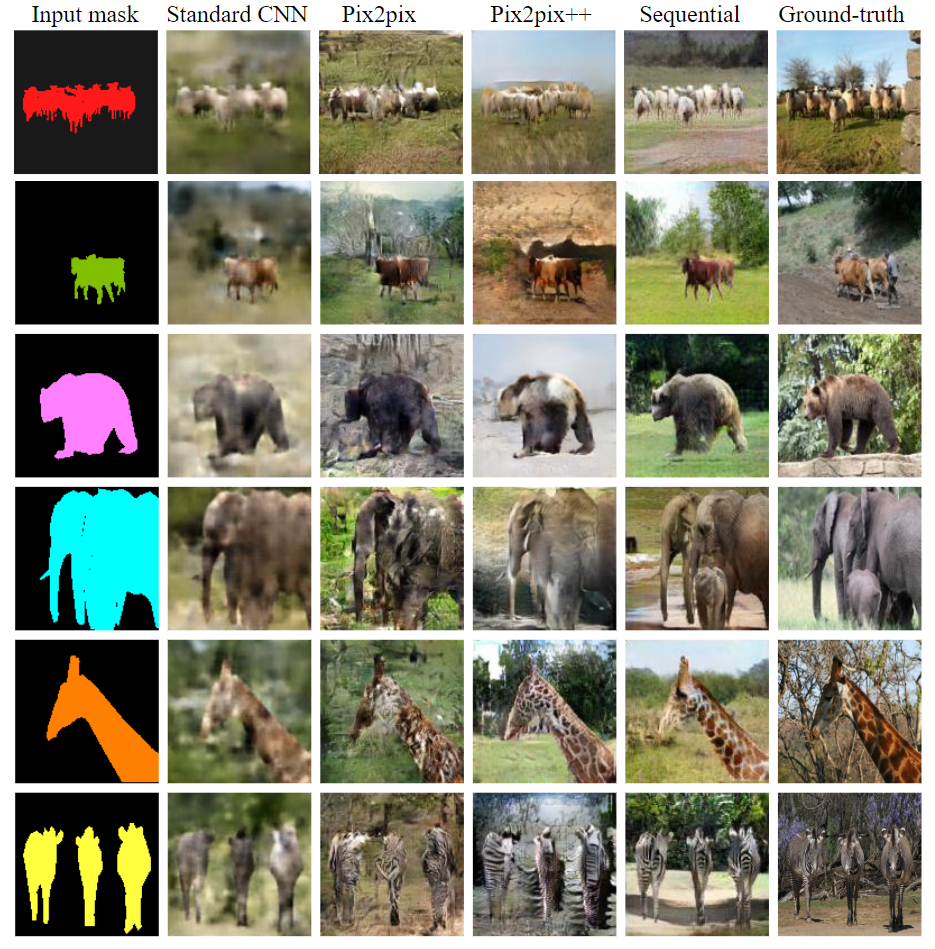}
\caption{Comparison with state of the art models using object masks from the validation set. From top to bottom: \textit{sheep}, \textit{cow}, \textit{bear}, \textit{elephant}, \textit{giraffe} and \textit{zebra}. The ground truth corresponds to the original image.
}
\label{val}
\end{figure}
\section{Results}

\subsection{Comparison with state of the art}
Figure~\ref{train} compares the visual result of the different baselines for the scene generation task for the six object classes in the dataset. The standard CNN produces blurry results despite an L1 loss to remediate this particular issue. Pix2Pix generates sharper images. However, both the standard CNN and Pix2Pix try to replicate the ground truth, which shows that it only memorizes the dataset. They lack stochasticity in the generation process, which dampens the diversity in the generated images. This behaviour should be avoided because the objective is to produce realistic scenes, not to mimic the scenes in the dataset. Diversity and image quality improve when adding the latest GANs training techniques as showcased in Pix2Pix++ and our proposed model. However, Pix2Pix++ seems to struggle when multiple objects are present (row 1, 2, 4 and 5). It cannot produce a plausible scene. Our model explicitly separate the foreground and background generation process, which overcomes these issues.

Figure~\ref{val} depicts samples generated using masks from the validation set. Compared to Figure~\ref{train}, it shows the generalization power of the different baseline models. The issues encountered by the respective baselines in Figure~\ref{train} are emphasized in Figure~\ref{val}. The standard CNN and Pix2Pix tend to produce an uniform background whereas the other two methods usually add a sky or trees besides the ground. Similarly, our proposed sequential model better handles multiple objects in a scene. The elephant sample (row 4) is an example. Our sequential model distinguishes three elephants while Pix2Pix++ only outputs one non-realistic object. 

\begin{table}[]
\centering
\begin{tabular}{@{}ll@{}}
\toprule
\textbf{Method}                & \textbf{FID}           \\ \midrule
Standard CNN                   & 120.7         \\
Pix2Pix                        & 34.0          \\
Pix2Pix++                      & 24.0          \\ \midrule
Sequential (proposed)          & 28.7          \\
Sequential (bg from Pix2pix++) & \textbf{23.2} \\ \bottomrule
\end{tabular}
\caption{Frechet Inception distance (lower the better). Our proposed model with background swapping outperforms the other baselines.}
\label{table1}
\end{table}

\begin{table}[]
\centering
\begin{tabular}{@{}lcc@{}}
\toprule
\textbf{Method}     & \multicolumn{2}{c}{\textbf{Mean IoU}}       \\
\cmidrule(lr){2-3}
                    & \textit{Train}          & \textit{Val.}            \\ \midrule
Standard CNN          & 0.298          & 0.272          \\
Pix2Pix               & 0.504          & 0.480          \\
Pix2Pix++             & 0.608          & 0.605          \\ \midrule
Sequential (proposed) & \textbf{0.650} & \textbf{0.650} \\ \midrule
Ground truth          & 0.803          & 0.770          \\ \bottomrule
\end{tabular}
\caption{Semantic segmentation accuracy (higher the better). The ground truth score is computed on the original images in the dataset and acts as an upper bound. Our proposed sequential model outperforms the other baselines.
}
\label{table2}
\end{table}

Table~\ref{table1} and Table~\ref{table2} support these claims with quantitative evaluations.
First, Table~\ref{table1} presents FID scores for the baseline models and the proposed model.
The standard CNN achieves a very bad score, which confirms the bad quality of the generated images.
Pix2Pix achieves a better score given the much sharper generated images.
Interestingly, Pix2pix++ yield a better FID score than our proposed model, 24.0 \textit{vs.} 28.7. After investigation, we found that outliers might arise when there is for example a sheep flying in the sky. In our current framework, a flying sheep is considered as a realistic scene because it gives a lot of freedom to the user to compose the scene. However, the FID score considers a flying sheep as a strong outlier. For a fair comparison, we swap the generated background of our proposed model with the generated background of Pix2Pix++. After the swapping, the FID score drops and is lower than Pix2Pix++, which shows that the foreground objects are better generated. 
Second, Table~\ref{table2} presents the mean IoU scores. In this experiment, an off-the-shelf model~\cite{deeplab} segments the generated images.
The ground truth (last row) corresponds to the mean IoU scores of the original images in the training or validation set. It then acts as an upper bound.
Adding an adversarial loss improves significantly the mean IoU score. Indeed, there is a jump of 0.3 point from the standard CNN to Pix2Pix.
Adding the latest GANs training techniques also improves the mean IoU score. It corresponds to an increase of 0.1 point from the original Pix2Pix to the enhanced Pix2Pix++.
The best scores are achieved by the proposed model.
Overall, the proposed sequential model generates images more realistically than the conventional models.

\begin{figure}[t]
\centering
\includegraphics[width=8.0cm]{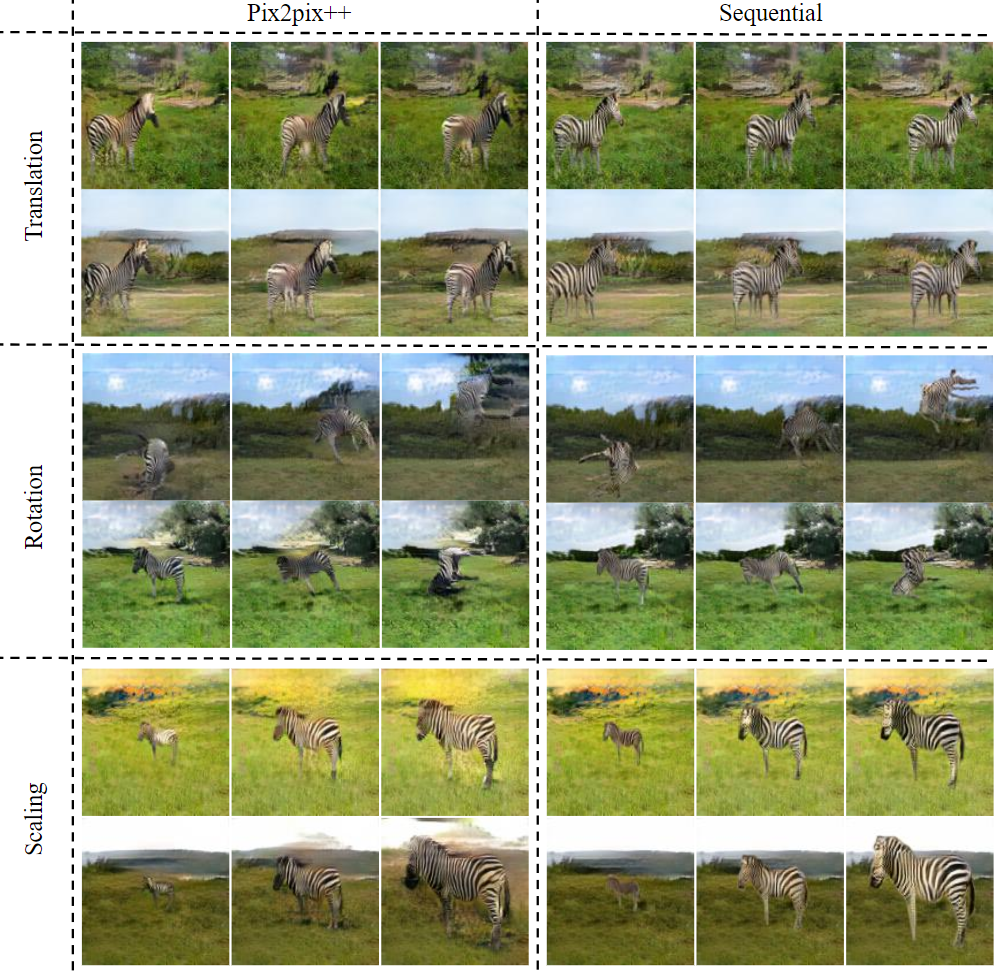}
\caption{Object mask affine transformations. Pix2Pix++ cannot preserve the scene. It either changes the background, blend the foreground and background or suffer from color blending. Our model does not suffer from these artifacts.
}
\label{affine}
\end{figure}

\subsection{Foreground object mask transformation}
In this experiment, several affine transformations are applied to the input object mask.
Figure~\ref{affine} shows how the Pix2pix++ and the proposed sequential model preserve the scene when translation, rotation and scaling operations transform the foreground object.
Note that a similar noise vector conditions the generation process for each subset of images.
Consider the Pix2Pix++ model (left column of Fig.~\ref{affine}).
When a translation is applied, both the foreground and the background are altered.
When the object is rotated, the foreground object starts to blend with the background. Pix2Pix++ seems to learn the color correlation between the ground and the legs of the zebra, which makes it impossible to draw objects with large rotations.
When the object is scaled up, colors of the foreground object start to bleed in the background.
In contrast, our proposed sequential model  (right column of Fig.~\ref{affine}) preserves the background up to affine transformations and do not suffer from object blending or color bleeding. It does however slightly change the appearance of the foreground objects.

\begin{figure}[t]
\centering
\includegraphics[width=8cm]{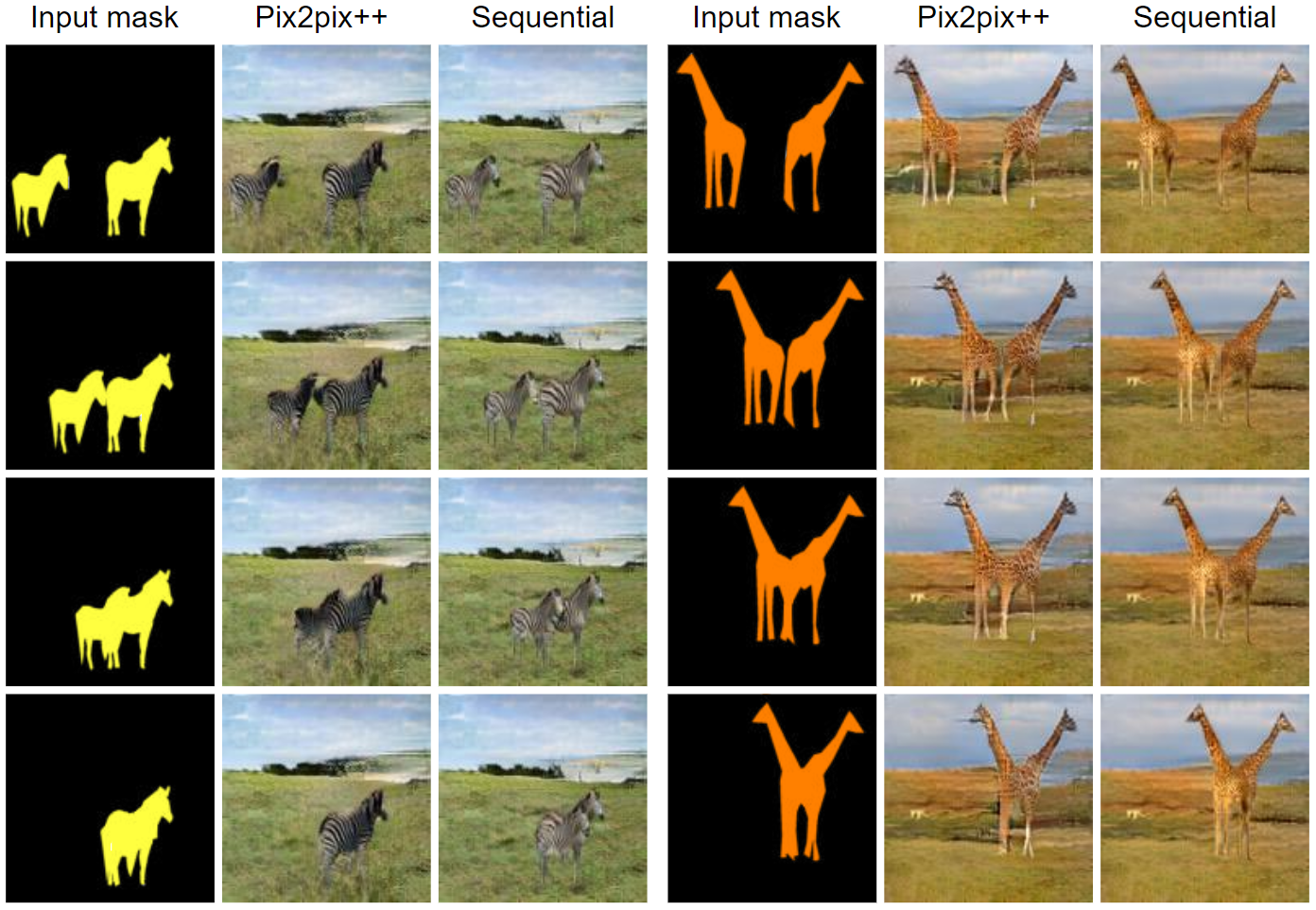}
\caption{Horizontal translation of two objects until they occlude each other. Pix2pix++ depicts artifacts on the foreground objects that are not present in our proposed model. }
\label{occlusion}
\end{figure}

\subsection{Foreground object occlusion}
In this experiment, two foreground objects are translated on the horizontal axis until they occlude each other.
Figure~\ref{occlusion} compares how the Pix2Pix++ and the proposed sequential model generate the foreground objects when their respective object masks get closer.
First, Pix2Pix++  cannot properly delineate the two objects.
Consider the zebra example and the Pix2pix++ model.
When the masks touch each other slightly (row 2), it splits the zebras at the wrong place.
When they are fully occluded (row 4), it only draws one single zebra.
Second, Pix2Pix++  tends to produce a similar pattern for both objects.
Consider now the giraffe example and the Pix2pix++ model.
When the masks get closer (row 2), it outputs similar colors for the giraffes.
When they touch each other slightly (row 3), it merges giraffes and draws a continuous pattern.
When they are fully occluded (row 4), it draws a giraffe with two heads.
These two types of artifact do not occur in our proposed sequential model.

 \subsection{Foreground objects control}
\begin{figure}[t]
 \centering
 \includegraphics[width=8cm]{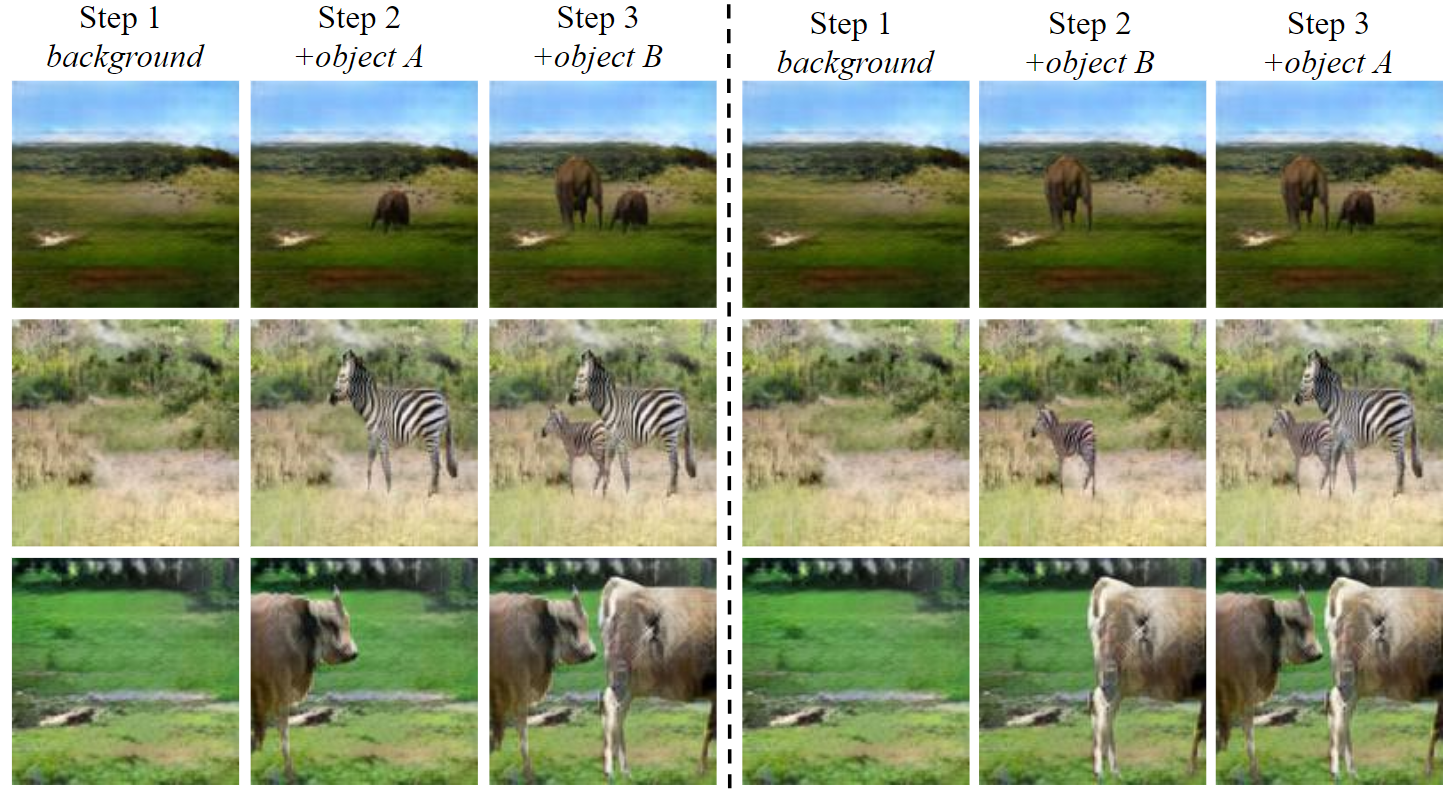}
 \caption{Composing a scene with a different order. The first order is on the left while the second order is on the right.}
 \label{order}
 \end{figure}

\begin{figure}[t]
 \centering
 \includegraphics[width=8cm]{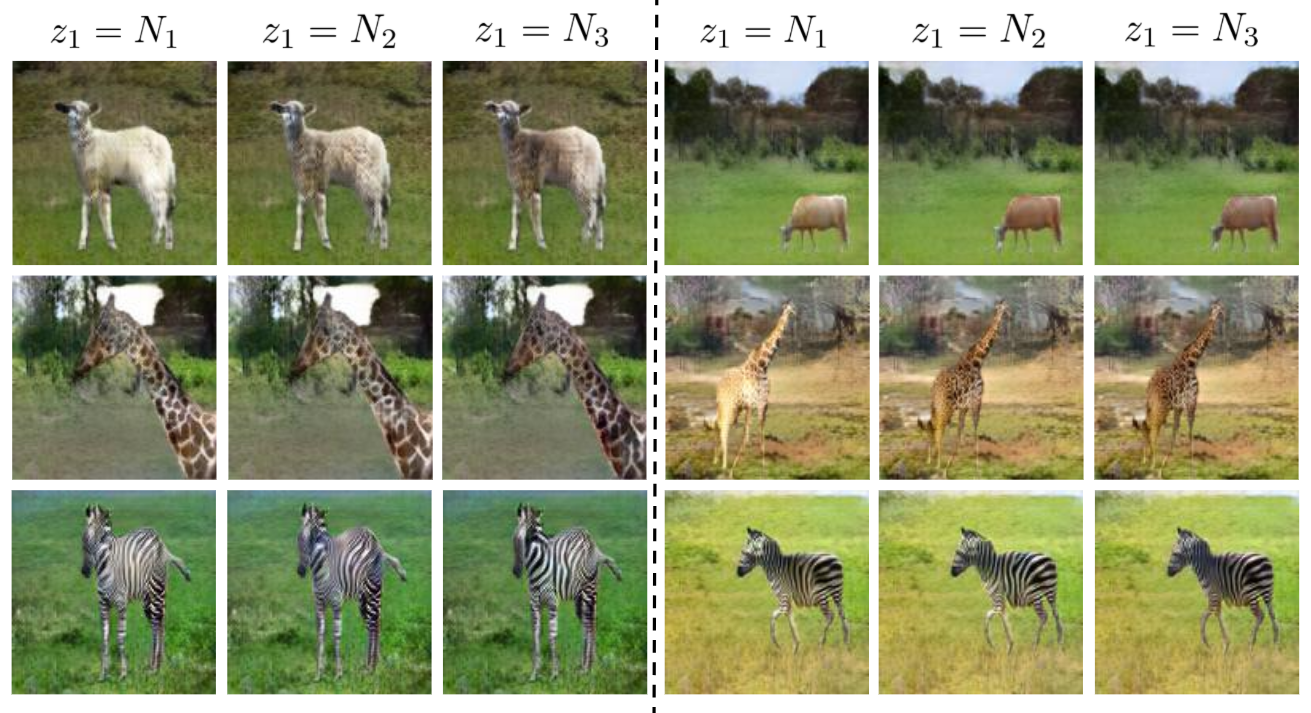}
 \caption{Controlling the appearance of the foreground object by altering associated input noise, $z_1$.}
 \label{var}
 \end{figure}

 First, we investigate whether object ordering makes difference in the final scene. In Figure \ref{order}, we generate the same scenes twice but with a different ordering of the object masks. We noticed small differences in terms of appearance,~\textit{e.g.} the order can have an influence on the illumination of the object.
 Secondly, we show that the object appearance can be altered by varying the associated noise. In Figure \ref{var}, we add a foreground object from different noise vectors but with the same mask on the same background. 
 
\subsection{Beyond generated background}
\begin{figure}[t]
 \centering
 \includegraphics[width=8cm]{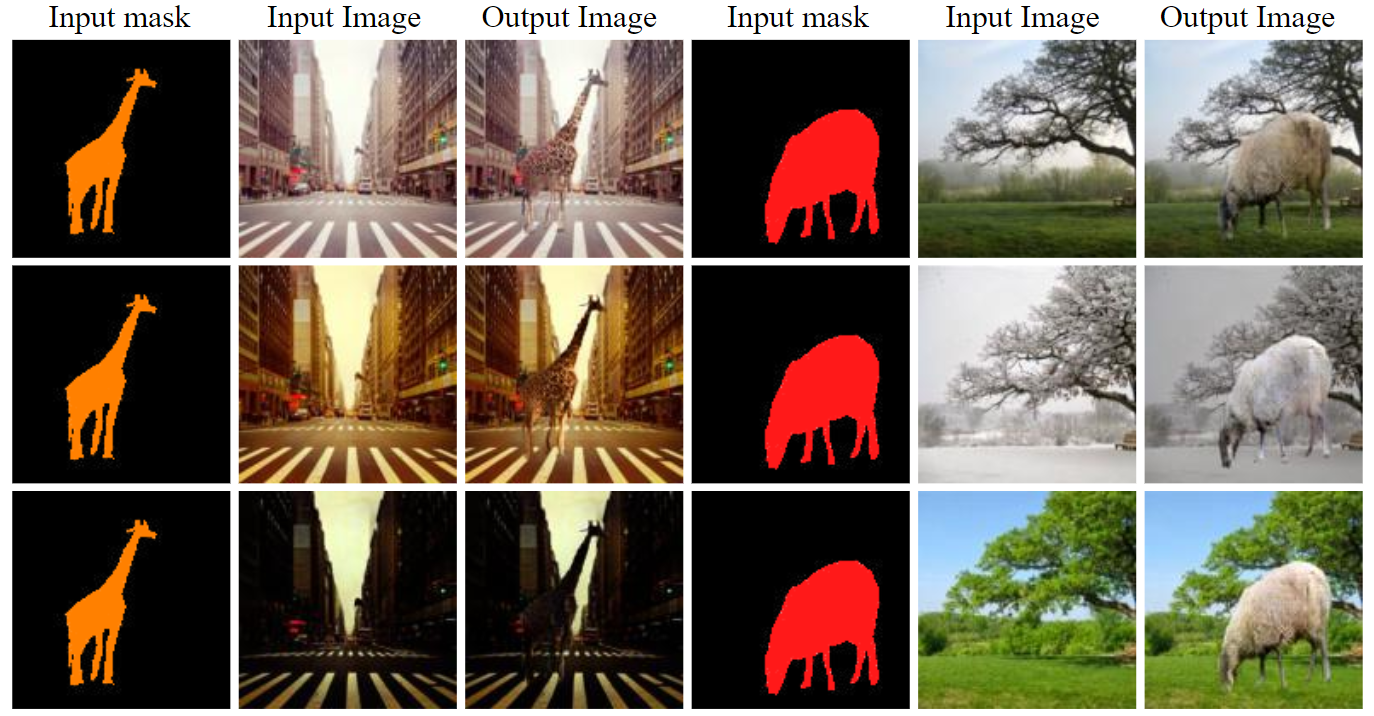}
 \caption{Adding new objects on existing images. Global illumination influences the object appearance.
 }
 \label{edit}
 \end{figure}

Our foreground model can also be used for image editing purposes by adding an object to an existing image. In Figure \ref{edit}, we add objects to the same scene but with different lighting or seasonal conditions. The foreground model is aware of the background scene in terms of its content and its environmental conditions such as global illumination.

\subsection{Beyond object masks}
\begin{figure}[t]
\centering
\includegraphics[width=8.0cm]{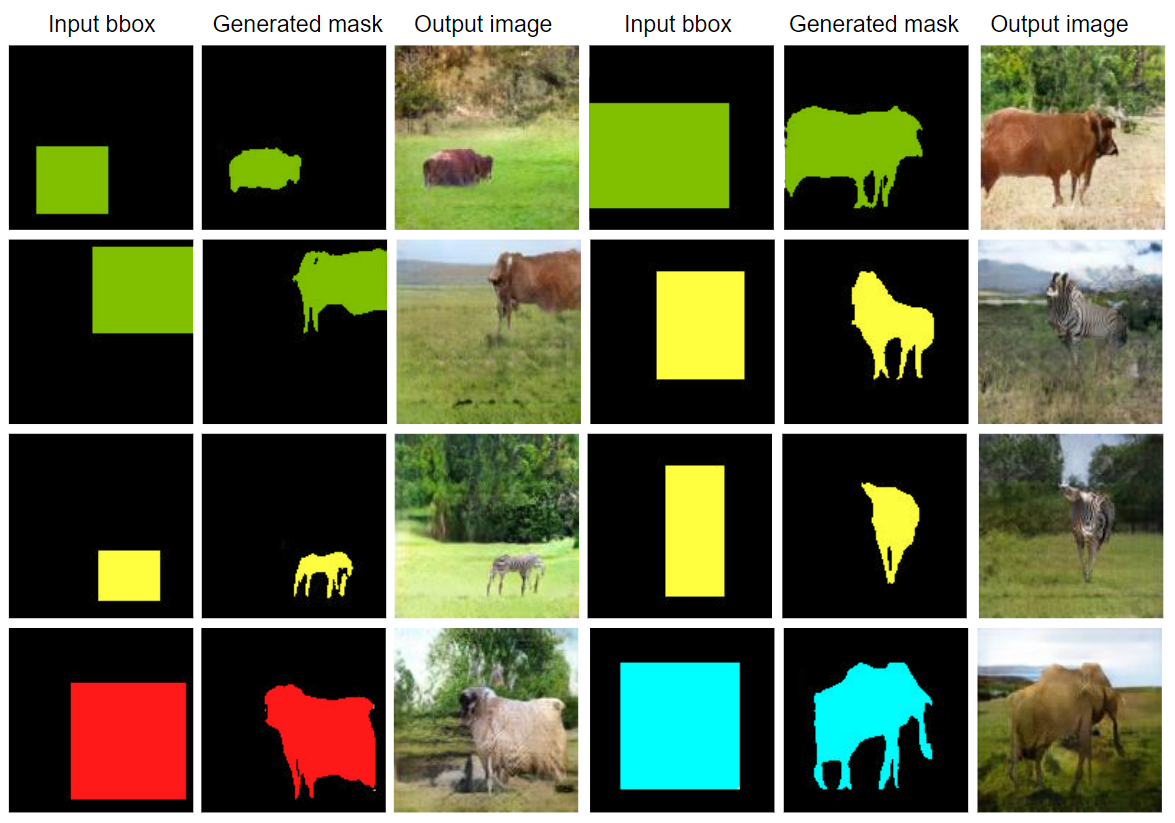}
\caption{Generating an image starting from a bounding box instead of an object mask. Note that the masks are generated in this experiment. The location of boxes controls the horizon line while the shape controls the viewpoint.
}
\label{mask}
\end{figure}

In the current problem formulation, the foreground generator is conditioned on object masks. This limits the user control over the proposed sequential model because a user would have to draw the shape of objects to obtain a scene.
In this experiment, we show how the current framework can easily be extended by introducing a separate mask generator model. The idea is to generate object shapes based on a bounding box. They can further be used as input to the current framework.
The mask generator is a conditional GANs model which takes a bounding box and object class as input and outputs an object mask in the region of interest. In order to stabilize the training, a cross-entropy reconstruction loss is added to the objective function. 
Figure~\ref{mask} presents generated mask samples from a bounding box and their final generated images. The locality of the bounding boxes influences over the horizon line. Boxes in the top (\textit{e.g.} row 2 left) generate a high horizon line while boxes in the bottom (\textit{e.g.} row 3 left) generate a low horizon line. The shape of the bounding boxes influences the viewpoint. Vertical rectangles (\textit{e.g.} row 2 right) generate a front view while squares or horizontal rectangles generate a side view.

\section{Conclusion}
In this paper, a sequential scene generation model based on Generative Adversarial Networks is proposed. This model adopts the layered structure modeling for images and generates an image step-by-step starting with the background of the scene and forms the scene progressively by drawing a single foreground object at each step. The proposed approach improves the controllability of the image generation process through a object-level control mechanism. The experimental results suggest that the sequential generation scheme also improves the image quality and the diversity. Finally, it is shown that it resolves the occlusion artifacts of the existing conditional GANs models.

\subsubsection{Acknowledgments.} We thank the anonymous reviewers for their valuable input. WT is partially supported by an NSERC scholarship.

\bibliographystyle{aaai19}
\bibliography{main}

\end{document}